\documentclass[conference]{IEEEtran}
\IEEEoverridecommandlockouts
\usepackage{cite}
\usepackage{amsmath,amssymb,amsfonts}
\usepackage{algorithmic}
\usepackage{graphicx}
\usepackage{textcomp}
\usepackage{xcolor}

\usepackage{algorithm}
\usepackage{booktabs}
\usepackage{multirow}
\usepackage{authblk}

\def\BibTeX{{\rm B\kern-.05em{\sc i\kern-.025em b}\kern-.08em
    T\kern-.1667em\lower.7ex\hbox{E}\kern-.125emX}}
\begin{document}

\title{Advancing Video Quality Assessment for AIGC}

\author[1]{Xinli Yue$^{*, }$\thanks{$^{*}$Equal contribution. Work done during Xinli Yue’s internship
at WeChat.}}
\author[2]{Jianhui Sun$^{*, }$}
\author[2]{Han Kong$^{*, }$}
\author[2]{Liangchao Yao}
\author[2]{Tianyi Wang}
\author[2]{Lei Li}
\author[2]{\\ Fengyun Rao}
\author[2]{Jing Lv}
\author[2]{Fan Xia}
\author[2]{Yuetang Deng}
\author[1]{Qian Wang}
\author[1]{Lingchen Zhao$^{\dag, }$\thanks{$^{\dag}$Corresponding author.}}
\affil[1]{School of Cyber Science and Engineering, Wuhan University, Wuhan, China}
\affil[2]{WeChat, Tencent Inc, Guangzhou, China}

\maketitle

\begin{abstract}
In recent years, AI generative models have made remarkable progress across various domains, including text generation, image generation, and video generation. However, assessing the quality of text-to-video generation is still in its infancy, and existing evaluation frameworks fall short when compared to those for natural videos. Current video quality assessment (VQA) methods primarily focus on evaluating the overall quality of natural videos and fail to adequately account for the substantial quality discrepancies between frames in generated videos. To address this issue, we propose a novel loss function that combines mean absolute error with cross-entropy loss to mitigate inter-frame quality inconsistencies. Additionally, we introduce the innovative S\textsuperscript{2}CNet technique to retain critical content, while leveraging adversarial training to enhance the model's generalization capabilities. Experimental results demonstrate that our method outperforms existing VQA techniques on the AIGC Video dataset, surpassing the previous state-of-the-art by 3.1\% in terms of PLCC.
\end{abstract}


\section{Introduction}
The rapid development of AI generative models~\cite{gan, auto, gpt4} has fueled advancements across various tasks, from text generation (Text-to-Text)~\cite{s2s, brown2020language} to image generation (Text-to-Image)~\cite{brock2018large, karras2019style, ramesh2021zero}, and more recently, video generation (Text-to-Video)~\cite{vondrick2016generating, wu2023tune, ho2022video}. Text-to-Text and Text-to-Image models have already achieved significant success in various applications, with extensive research and mature evaluation methods backing their progress~\cite{bleu, lin2004rouge, zhang2019bertscore, zhang2023perceptual, yu2024sf, peng2024aigc, yuan2024tier}. However, compared to these domains, the task of Text-to-Video generation is more complex and challenging, and the methods for evaluating its output quality remain underdeveloped. Current research on Text-to-Video evaluation is relatively scarce, underscoring the urgent need for more exploration in this area. Developing robust and reliable evaluation methods is crucial to establishing a solid theoretical foundation and offering practical guidance for the advancement of future generative models.

Numerous studies have focused on quality assessment for natural videos. For instance, VSFA~\cite{vsfa} leverages deep neural networks to perform no-reference video quality assessment by integrating content-dependence and temporal memory effects. SimpleVQA~\cite{simplevqa} trains an end-to-end multi-scale spatial feature extractor and uses a pre-trained, fixed motion extractor to capture features for quality regression. FAST-VQA~\cite{fastvqa} utilizes grid-based patch sampling and a fragment attention network to efficiently and accurately assess the quality of high-resolution videos, significantly reducing computational costs. Building on FAST-VQA~\cite{fastvqa}, SAMA~\cite{sama} enhances the performance of single-branch models by using a scaling and masking sampling strategy, compressing both local and global content into standard input sizes.

Due to the relatively small inter-frame quality variations in natural videos, most prior works~\cite{vsfa,simplevqa,fastvqa,sama} focus on assessing video quality as a whole. However, with current technical limitations, AIGC videos exhibit significantly larger inter-frame quality variations compared to natural videos, where some frames are of high quality while others are of lower quality. If we directly apply the mean absolute error (MAE) loss~\cite{simplevqa} between the subjective video score and the mean of the predicted frame-wise scores, the model may fail to effectively capture the quality fluctuations between frames, potentially losing critical information. Alternatively, using a binary cross-entropy (BCE) loss between the true video score distribution and the predicted per-frame score distribution penalizes videos with the same mean predicted score differently based on inter-frame variations. For example, video A (with frame-wise predicted scores of 1, 2, 3) would be penalized more than video B (with frame-wise predicted scores of 2, 2, 2), despite having the same mean score, which is evidently unfair.

To address the issue of inter-frame quality variations in AIGC video quality assessment (VQA) tasks, we propose a novel loss function, \textbf{F}rame \textbf{C}onsistency \textbf{L}oss (\textbf{FCL}). FCL is defined as the product of the MAE loss between the subjective video score and the mean predicted frame-wise scores, and the BCE loss between the true video score distribution and the predicted frame-wise score distribution. This formulation not only stabilizes training and mitigates overfitting but also alleviates the problem of inter-frame quality discrepancies.

Moreover, in previous VQA work~\cite{simplevqa, fastvqa, sama}, video frame sampling methods have primarily relied on random cropping or grid-based patch sampling. However, these approaches risk losing crucial content, potentially omitting essential information. To address this, we propose a novel sampling strategy using S\textsuperscript{2}CNet~\cite{wecrop}, which performs content-aware cropping to preserve important regions, thereby capturing richer and more comprehensive features.

Additionally, adversarial training~\cite{pgd}, initially introduced to enhance adversarial robustness in image classification, often degrades performance on clean samples~\cite{trades}. Interestingly, recent work~\cite{fgm} has demonstrated that applying adversarial training in text classification can actually improve generalization on clean samples. This raises curiosity about its impact on VQA tasks. Motivated by this, we explore the application of adversarial training in VQA, specifically by introducing adversarial perturbations to the model weights using Fast Gradient Method (FGM)~\cite{fgm} and optimizing the model accordingly.



\begin{figure*}[htbp]
    \centering
    \includegraphics[width=0.8\textwidth]{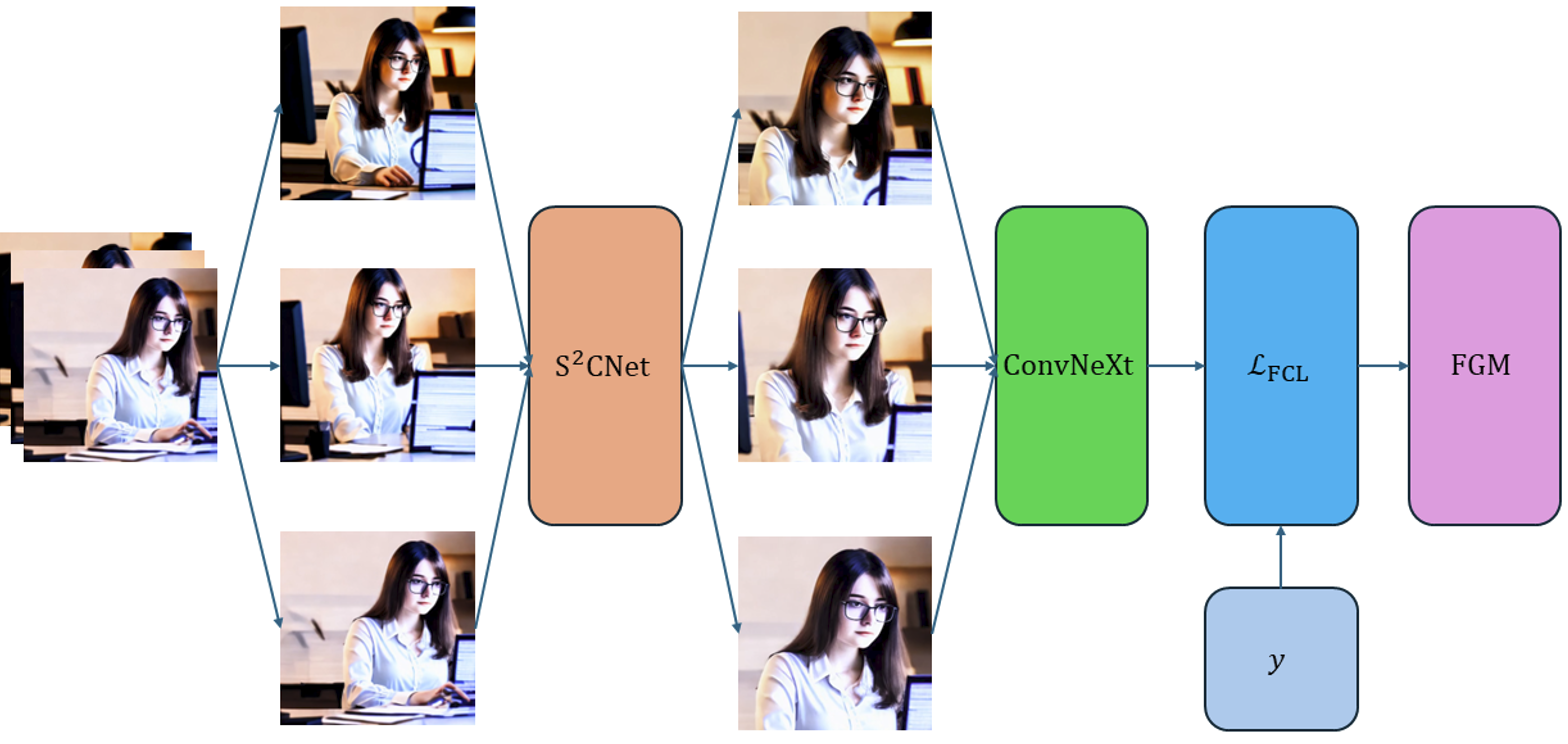}
    \caption{Overview}
    \label{fig_overview}
\end{figure*}

\section{Methodology}

\subsection{Frame Consistency Loss}

\paragraph{MAE Loss} For a classic regression model, the training objective is to minimize the mean absolute error (MAE) between the target video score and the mean of the predicted scores for each frame (or video segment)~\cite{simplevqa}:

\begin{equation}
    \mathcal{L}_{\mathrm{MAE}}=\left|\frac{1}{F} \sum_{f=1}^F \hat{y}_f^{\mathrm{frame}}-y\right|
\end{equation}
where $\hat{y}_f^{\text {frame }}$ is the predicted score for the $f$-th frame of the video sample, $F$ is the number of sampled frames, and $y$ is the true quality score of the video. However, using MAE loss may cause the model to fail in capturing inter-frame quality variations effectively, potentially losing important information.

\paragraph{BCE Loss} We first generate a score sequence ranging from 0 to 99 , denoted by $s_i$ as the $i$-th score:
\begin{equation}
    s_i=i, \quad i=0,1,2, \ldots, 99
\end{equation}

The model outputs a predicted probability distribution with shape $[F, 100]$, where $F$ is the number of frames and 100 is the number of score categories. To obtain the predicted score for each frame, we compute a weighted average between the frame's probability distribution and the score vector:
\begin{equation}
    \hat{y}_f^{\text {frame }}=\frac{\sum_{i=0}^{99} p_{f, i} \cdot s_i}{\sum_{i=0}^{99} p_{f, i} \cdot 100}
\end{equation}
where $\hat{y}_f^{\text {frame }}$ is the predicted score for the $f$-th frame, and $p_{f, i}$ is the model's predicted probability for the $i$-th score of the $f$-th frame.

The frame-level BCE loss is computed as follows:
\begin{equation}
    \mathcal{L}_{\mathrm{BCE}}=\frac{1}{F \cdot 100} \sum_{f=1}^F \sum_{i=0}^{99} \operatorname{BCE}\left(d_{f, i}, p_{f, i}\right)
\end{equation}

\begin{equation}
    d_{f, i}=\frac{1}{\sigma \sqrt{2 \pi}} \exp \left(-\frac{\left(s_i-y\right)^2}{2 \sigma^2}\right)
\end{equation}
where, $d_{f, i}$ represents the Gaussian-distributed label for each frame of the video sample, and $\sigma$ is the standard deviation of the ground-truth video scores across the dataset.

\paragraph{Frame Consistency Loss}
Compared to MAE loss, BCE loss can leverage more information about inter-frame quality variations. However, AIGC videos often exhibit larger inter-frame quality disparities than natural videos. For example, video A (with a human rating of 2) may have three frames with predicted scores of 1, 2, and 3, respectively, while video B (also rated 2) may have three frames, all predicted as 2. If only BCE loss is used, video A and video B would be penalized differently, which would be unfair to video A.

To address this issue, we propose a novel loss function, Frame Consistency Loss (FCL), defined as:

\begin{equation}
    \mathcal{L}_{\mathrm{FCL}}=\mathcal{L}_{\mathrm{MAE}} \times \mathcal{L}_{\mathrm{BCE}}
\end{equation}

In this formulation, when the mean of the predicted frame scores equals the ground-truth video score, $\mathcal{L}_{\mathrm{MAE}}$ becomes zero, and consequently, $\mathcal{L}_{\mathrm{FCL}}$ also becomes zero. This ensures fairness for videos like A with larger inter-frame quality variations. 

\subsection{S\textsuperscript{2}CNet}
Previous works~\cite{simplevqa, fastvqa} typically employed random cropping or grid-based patch sampling for video frame extraction. However, these methods risk losing key parts of the image, resulting in incomplete or suboptimal content representation. Human annotators, when scoring videos, tend to focus on the most visually salient regions. Thus, we propose using an intelligent cropping algorithm, S\textsuperscript{2}CNet~\cite{wecrop}, to enhance the aesthetic quality and content preservation of the cropped video frames.

Specifically, following ~\cite{wecrop}, given an input video frame $I$ and its associated candidate cropping regions, we employ a Faster R-CNN~\cite{fastrcnn} pre-trained on VisualGenome~\cite{krishna2017visual} to identify the top $N$ potential visual objects. Next, the image is passed through a convolutional backbone~\cite{simonyan2014very, sandler2018mobilenetv2} to obtain feature maps $F$. We then apply RoIAlign~\cite{he2017mask} and RoDAlign~\cite{gao2019graph} operations, followed by a fully connected layer, to extract $d$-dimensional features from the overall visual regions, denoted as $\left[x_1, x_2, \ldots, x_{N+1}\right] \in \mathbb{R}^{(N+1) \times d}$ (representing $N$ detected objects and one cropping candidate region). Next, these features are fed into the S\textsuperscript{2}CNet network to capture higher-order information for aesthetic scoring. The structure of S\textsuperscript{2}CNet is as follows: 

\paragraph{Adaptive Attention Map}We first establish the semantic relationships between nodes by encoding pairwise relations and assigning different weights to edges. To do this, we compute the appearance similarity matrix $\mathcal{M}_a \in \mathbb{R}^{(N+1) \times(N+1)}$ in the embedding space to capture feature correlations, as shown below:
\begin{equation}
    \mathcal{M}_{a(i, j)}=\frac{\phi\left(x_i\right)^T \varphi\left(x_j\right)}{\sqrt{d}}
\end{equation}

where $\phi(x)=W_\phi x+b_\phi$ and $\varphi(x)=W_{\varphi} x+b_{\varphi}$ are two learnable linear functions.

Next, we establish spatial information between nodes. The center coordinates of the bounding box of node $x_i$, denoted as $p_i=\left(p_i^x, p_i^y\right)$, serve as the initial spatial features. We explicitly model the spatial connections between nodes and represent the spatial position matrix $\mathcal{M}_p$ as:
\begin{equation}
    \mathcal{M}_{p(i, j)}=\left\|\left(W_m p_i+b_m\right)-\left(W_n p_j+b_n\right)\right\|_2^2
\end{equation}

where $W_{m ; n}$ and $b_{m ; n}$ are different learnable weight matrices and biases.

To jointly capture sufficient spatial-semantic information, we construct the spatial-semantic adjacency matrix $\mathcal{A} \in \mathbb{R}^{(N+1) \times(N+1)}$ as a combination of the following form:
\begin{equation}
    \mathcal{A}_{(i, j)}=\frac{\mathcal{M}_{a(i, j)} \cdot e^{\mathcal{M}_{p(i, j)}}}{\sum_{j=1}^{N+1} \mathcal{M}_{a(i, j)} \cdot e^{\mathcal{M}_{p(i, j)}}}
\end{equation}

\paragraph{Graph-Aware Attention Module} Once the graph is assembled, feature extraction is performed on the nodes. We adopt a Transformer-like graph-aware attention operation, but combine spatial and semantic features to generate attention weights. Before applying self-attention to the node features, we pass them through a feature aggregation gate (FAG) to implicitly embed the adjacency tensor information. Specifically, treating the nodes as tokens, and given the input features X and the corresponding adjacency tensor A, the computation of FAG is as follows:
\begin{equation}
    X=\operatorname{RELU}(\mathcal{A} Z X)
\end{equation}

where $Z \in \mathbb{R}^{(N+1) \times d}$ is a learnable weight matrix. The output feature $X$ aggregates information from neighboring nodes.

Next, the output of FAG is treated as the query $Q$, while the original node features are used as the key $Q$ and value $V$. The self-attention mechanism is then redefined as follows:
\begin{equation}
    S^2 O-SA=\operatorname{softmax}\left(\frac{Q K^T}{\sqrt{d}}+\mathcal{M}_a+\mathcal{M}_p\right) V
\end{equation}

\paragraph{Score Regression} Finally, after obtaining the features from the adaptive attention map, we use a two-layer multilayer perceptron (MLP) to aggregate the updated information from all nodes to predict the aesthetic score. The score regression is performed using a weighted smooth $\ell_1$ loss and ranking loss.

The cropped candidate region with the highest aesthetic score is then selected as the final cropped image.

\subsection{Adversarial Training}
Adversarial training~\cite{pgd} is proposed as a defense mechanism against adversarial attacks, primarily targeting image classification. 
It has been noted that while adversarial training can enhance robustness, it often leads to a degradation in generalization performance on clean samples, highlighting a trade-off between natural generalization and robust generalization~\cite{trades}.
However, some studies~\cite{fgm} have proposed that applying adversarial perturbations to text embeddings can improve the generalization performance of models in text classification tasks. 

In image classification and text classification, adversarial training appears to have contradictory effects on clean generalization. 
This raises curiosity about the impact of adversarial training when applied to VQA tasks. Therefore, we attempt to introduce weight perturbations in the AIGC VQA task. Specifically, we need to determine the direction of the perturbation. Following FGM~\cite{fgm}, the perturbation direction is aligned with the gradient direction, i.e., the direction that maximizes the loss. The perturbation $\delta$ is computed as follows:
\begin{equation}
    \delta \leftarrow \epsilon \cdot \frac{\nabla_w \mathcal{L}}{\left\|\nabla_w \mathcal{L}\right\|}
\end{equation}

where $\mathcal{L}$ is the loss function, $w$ represents the weights, and $\epsilon$ is a hyperparameter controlling the perturbation magnitude. After computing the perturbation, we add it to the original weights:
\begin{equation}
    w^{\prime} \leftarrow w+\delta
\end{equation}

Then, we update the model parameters as follows:
\begin{equation}
    w \leftarrow w-\eta \cdot\left(\nabla_w \mathcal{L}+\nabla_{w^{\prime}} \mathcal{L}_{a d v}\right)
\end{equation}

where $\mathcal{L}_{a d v}$ is the loss calculated using the perturbed weights $w^{\prime}$, and $\eta$ is the learning rate. The FGM algorithm is outlined in Algorithm \ref{alg_fgm}.

\begin{algorithm}[t]
\caption{FGM}
\label{alg_fgm}
\textbf{Input}: model $\mathbf{f_w}$, batch size $m$, learning rate $\eta$,  perturbation size $\epsilon$ \\
\textbf{Output}: model $\mathbf{f_w}$
\begin{algorithmic}[1] 
\REPEAT
\STATE Read mini-batch $B={\mathbf{x}_1,\ldots,\mathbf{x}_m }$ from training set
\STATE Compute loss $\mathcal{L}$ on $B$
\STATE Compute gradients $\nabla_w \mathcal{L}$
\STATE $\quad \delta \leftarrow \epsilon \cdot \frac{\nabla_w \mathcal{L}}{\left\|\nabla_w \mathcal{L}\right\|}$
\STATE $\quad w^{\prime} \leftarrow w+\delta$
\STATE Compute adversarial loss $\mathcal{L}_{a d v}$ on $B$
\STATE Compute gradients $\nabla_{w^{\prime}} \mathcal{L}_{a d v}$
\STATE $w \leftarrow w-\eta \cdot\left(\nabla_w \mathcal{L}+\nabla_{w^{\prime}} \mathcal{L}_{a d v}\right)$
\UNTIL{training converged}
\end{algorithmic}
\end{algorithm}

\section{Experiments}
\subsection{Experimental Setup}
\paragraph{Implementation Details} We employ ConvNeXt~\cite{liu2022convnet} as the backbone network for the feature extraction module. 

\paragraph{Evaluation Metrics} We adopt three standard metrics to evaluate the performance of VQA models: Pearson Linear Correlation Coefficient (PLCC), Spearman Rank-Order Correlation Coefficient (SROCC), and Kendall Rank-Order Correlation Coefficient (KROCC). 


\paragraph{Comparison Methods} We compare our method against four baseline approaches: VSFA~\cite{vsfa}, SimpleVQA~\cite{simplevqa}, FAST-VQA~\cite{fastvqa}, and SAMA~\cite{sama}. 

\subsection{Experimental Results}
Table \ref{tab_main} presents the performance comparison between baseline methods~\cite{vsfa, simplevqa, fastvqa, sama} and our proposed approach on the T2VQA-DB dataset~\cite{T2VQA-DB}. As shown, our method achieves the best performance in all metrics, outperforming the second-best method SAMA~\cite{sama} by 2.8\%. This demonstrates the superior effectiveness of our approach for the AIGC VQA task.

\begin{table}[htbp]
\caption{Performance metrics of various algorithms on the T2VQA-DB dataset~\cite{T2VQA-DB}}
\begin{center}
\setlength{\tabcolsep}{4mm}
{
\begin{tabular}{@{}ccccc@{}}
\toprule[1.0pt]
Method    & SROCC          & PLCC           & KROCC          & Mean           \\ \midrule
VSFA~\cite{vsfa}      & 0.671          & 0.687          & 0.485          & 0.614          \\
SimpleVQA~\cite{simplevqa} & 0.638          & 0.650          & 0.458          & 0.582          \\
FAST-VQA~\cite{fastvqa}  & 0.705          & 0.722          & 0.521          & 0.649          \\
SAMA~\cite{sama}      & 0.713          & 0.726          & 0.528          & 0.656          \\ \midrule
Ours      & \textbf{0.742} & \textbf{0.757} & \textbf{0.555} & \textbf{0.684} \\ \bottomrule[1.0pt]
\end{tabular}
}
\label{tab_main}
\end{center}
\end{table}

\subsection{Ablation Study}
To investigate the contribution of the three key components in our proposed approach, we conducted a detailed ablation study. The results, presented in Table \ref{tab_ablation}, demonstrate that FCL, S\textsuperscript{2}CNet, and FGM each provide significant improvements for the AIGC VQA task. This highlights the importance of each component in enhancing the overall performance of the model.

\begin{table}[htbp]
\caption{Ablation study results of our proposed method on the T2VQA-DB dataset}
\begin{center}
\setlength{\tabcolsep}{2.8mm}
{
\begin{tabular}{@{}ccccccc@{}}
\toprule[1.0pt]
FCL & S\textsuperscript{2}CNet & FGM & SROCC          & PLCC           & KROCC          & Mean           \\ \midrule
    &        &     & 0.721          & 0.734          & 0.534          & 0.663          \\
\checkmark   &        &     & 0.727          & 0.742          & 0.541          & 0.670          \\
\checkmark   &        & \checkmark   & 0.730          & 0.744          & 0.544          & 0.673          \\
\checkmark   & \checkmark      &     & 0.740          & 0.753          & 0.553          & 0.682          \\
\checkmark   & \checkmark      & \checkmark   & \textbf{0.742} & \textbf{0.757} & \textbf{0.555} & \textbf{0.684} \\ \bottomrule[1.0pt]
\end{tabular}
}
\label{tab_ablation}
\end{center}
\end{table}

\subsection{Results in NTRIE 2024 S-UGC VQA}

In the NTRIE 2024 S-UGC VQA challenge~\cite{li2024ntire}, our FCL-based approach achieved the second-best result. This indicates that the proposed FCL also demonstrates strong evaluation capabilities for short-form videos, further validating its effectiveness across different video types.

\begin{table}[htbp]
\caption{Competition results of NTRIE 2024 S-UGC VQA, where we achieved the second-best performance}
\begin{center}
\setlength{\tabcolsep}{1.1mm}
{\begin{tabular}{@{}ccccccc@{}}
\toprule[1.0pt]
Rank                      & Team       & Final Score & SROCC  & PLCC   & Rank1  & Rank1  \\ \midrule
1                         & SJTU MMLab    & 0.9228      & 0.9361 & 0.9359 & 0.7792 & 0.8284 \\
\underline{2}                         & \underline{IH-VQA}        & \underline{0.9145}      & \underline{0.9298} & \underline{0.9325} & \underline{0.7013} & \underline{0.8284} \\
3                         & TVQE          & 0.9120      & 0.9268 & 0.9312 & 0.6883 & 0.8284 \\
4                         & BDVQAGroup    & 0.9116      & 0.9275 & 0.9211 & 0.7489 & 0.8462 \\
5                         & VideoFusion   & 0.8932      & 0.9026 & 0.9071 & 0.7186 & 0.8580 \\
6                         & MC2Lab        & 0.8855      & 0.8966 & 0.8977 & 0.7100 & 0.8521 \\
7                         & Padding       & 0.8690      & 0.8841 & 0.8839 & 0.6623 & 0.8047 \\
8                         & ysy0129       & 0.8655      & 0.8759 & 0.8777 & 0.6883 & 0.8402 \\
9                         & lizhibo       & 0.8641      & 0.8778 & 0.8822 & 0.6494 & 0.7929 \\
10                        & YongWu        & 0.8555      & 0.8629 & 0.8668 & 0.6970 & 0.8462 \\
11                        & we are a team & 0.8243      & 0.8387 & 0.8324 & 0.6234 & 0.8225 \\
12                        & dulan         & 0.8098      & 0.8164 & 0.8297 & 0.5758 & 0.8047 \\
13                        & D-H           & 0.7677      & 0.7774 & 0.7832 & 0.5931 & 0.7160 \\ \midrule
\multirow{3}{*}{Baseline} & VSFA~\cite{vsfa}          & 0.7869      & 0.7974 & 0.7950 & 0.6190 & 0.7870 \\
                          & SimpleVQA~\cite{simplevqa}     & 0.8159      & 0.8306 & 0.8202 & 0.6147 & 0.8461 \\
                          & FastVQA~\cite{fastvqa}       & 0.8356      & 0.8473 & 0.8467 & 0.6494 & 0.8166 \\ \bottomrule[1.0pt]
\end{tabular}
}
\label{tab_kvq}
\end{center}
\end{table}

\section{Conclusion}

Through a comprehensive analysis of existing video quality assessment methods and the unique challenges posed by AIGC videos, we proposed an innovative loss function and introduced an intelligent cropping strategy, along with adversarial training, into video quality assessment. The experimental results validate the advantages of our approach in addressing inter-frame quality discrepancies, significantly improving the overall performance of the model. In summary, our method offers an effective solution for AIGC video quality assessment.

\clearpage

\bibliographystyle{IEEEtran}
\bibliography{ref}

\end{document}